%% file: main.tex
\pdfoutput=1

\documentclass[11pt]{article}

\usepackage{emnlp2021}

\usepackage{times}
\usepackage{latexsym}

\usepackage[T1]{fontenc}

\usepackage[utf8]{inputenc}

\usepackage{microtype}

\usepackage{graphicx}
\usepackage{xcolor}
\usepackage{comment}
\usepackage{booktabs}
\usepackage{amsmath}
\usepackage{amssymb}
\usepackage{amsthm}
\usepackage{bbm} 
\usepackage{subcaption}
\usepackage{multirow}
\usepackage{comment}
\usepackage{makecell} 
\usepackage{tabularx}

\newcommand{\methodname}{{SYSML}} 
\usepackage[figuresleft]{rotating}

\theoremstyle{definition}
\newtheorem{definition}{Definition}[section]
\title{\methodname{}: {S}t{Y}lometry with {S}tructure and {M}ultitask {L}earning: Implications for Darknet Forum Migrant Analysis}

\author{Pranav Maneriker~\;~ Yuntian He~\;~Srinivasan Parthasarathy \\
The Ohio State University \\
\texttt{\{maneriker.1@,he.1773,parthasarathy.2\}@osu.edu}
}
\begin{document}

\maketitle
\begin{abstract}
Darknet market forums are frequently used to exchange illegal goods and services between parties who use encryption to conceal their identities. The Tor network is used to host these markets, which guarantees additional anonymization from IP and location tracking, making it challenging to link across malicious users using multiple accounts (sybils). Additionally, users migrate to new forums when one is closed further increasing the difficulty of linking users across multiple forums. We develop a novel stylometry-based multitask learning approach for natural language and model interactions using graph embeddings to construct low-dimensional representations of short episodes of user activity for authorship attribution. We provide a comprehensive evaluation of our methods across four different darknet forums demonstrating its efficacy over the state-of-the-art, with a lift of up to 2.5X on Mean Retrieval Rank and 2X on Recall@10.
\end{abstract}

\section{Introduction}
\input{sections/introduction}

\section{Related Work}
\input{sections/related}
\section{\methodname{} Framework}
\input{sections/methodology.tex}

\section{Datasets}
\input{sections/dataset.tex}

\section{Evaluation}
\input{sections/eval.tex}

\section{Analysis}
\input{sections/analysis.tex}

\section{Case Study}
\input{sections/casestudy.tex}

\section{Conclusion}
\input{sections/conclusion}

\section*{Acknowledgements}
\input{sections/ack}

\bibliography{anthology,custom}
\bibliographystyle{acl_natbib}

\appendix

\section{Ethics Statement}
\input{sections/ethics}

\input{sections/supplementary/supp}

\end{document}

%% file: sections/introduction.tex
Crypto markets are \textit{``online forums where goods and services are exchanged between parties who use digital encryption to conceal their identities''}~\cite{martin2014drugs}. 
  They are typically hosted on the Tor network, which guarantees  anonymization in terms of IP and location tracking. 
  The identity of individuals on a crypto-market is associated only with a username; therefore, building trust on these networks does not follow conventional models prevalent in eCommerce. 
  Interactions on these forums are facilitated by means of text posted by their users. 
  This makes the analysis of textual style on these forums a compelling problem.
  
  Stylometry is the branch of linguistics concerned with the analysis of authors' style.
  Text stylometry was initially popularized in the area of forensic linguistics, specifically to the problems of author profiling and author attribution~\cite{juola2008authorship, rangel2013overview}.
  Traditional techniques for authorship analysis on such data rely upon the existence of long text corpora from which features such as the frequency of words, capitalization, punctuation style, word and character n-grams, function word usage can be extracted and subsequently fed into any statistical or machine learning classification framework, acting as an author's `signature'. 
  However, such techniques find limited use in short text corpora in a heavily anonymized environment.

  Advancements in using neural networks for character and word-level modeling for authorship attribution aim to deal with the scarcity of easily identifiable `signature' features and have shown promising results on shorter text~\cite{shrestha-etal-2017-convolutional}. 
  \citet{andrews-witteveen-2019-unsupervised} drew upon these advances in stylometry to propose a model for building representations of social media users on Reddit and Twitter. Motivated by the success of such approaches, we develop a novel methodology for building authorship representations for posters on various darknet markets. 
  Specifically, our key contributions include: 
  
  \noindent \textbf{First}, a {\it representation learning} approach that couples temporal content stylometry with access identity (by levering forum interactions via \textit{meta-path graph context information}) to model and enhance user (author) representation; 

  \noindent \textbf {Second}, a novel framework for training the proposed models in a \textit{multitask setting} across multiple darknet markets, 
  using a small dataset of labeled migrations, 
  to refine the representations of users within each individual market, while also providing a method to correlate users across markets; 
  
\noindent \textbf{Third}, a detailed drill-down {\it ablation study} discussing the impact of various optimizations and highlighting the benefits of both graph context and multitask learning  on forums associated with four darknet markets - \textit{Black Market Reloaded}, \textit{Agora Marketplace}, \textit{Silk Road}, and \textit{Silk Road 2.0} -
 when compared to the state-of-the-art alternatives.

  
  

%% file: sections/related.tex

\noindent {\bf Darknet Market Analysis:}
Content on the dark web includes resources devoted to illicit drug trade, adult content, counterfeit goods and information, leaked data, fraud, and other illicit services~\cite{lapata2017proceedings,biryukov2014content} . 
Also included are forums discussing politics, anonymization, and cryptocurrency. 
\citet{biryukov2014content} found that while a vast majority of these services were in English (about $84\%$), a total of about 17 different languages were detected.
Analysis of the volume of transactions and number of users on darknet markets indicates that they are resilient to closures; rapid migrations to newer markets occur when one market shuts down~\cite{elbahrawy2019collective}.

Recent work~\cite{fan2018automatic,hou2017hindroid,fu2017hin2vec,dong2017metapath2vec} has levered the notion of a heterogeneous information network (HIN) embedding to improve graph modeling, where different types of nodes, relationships (edges) and paths can be represented through typed entities.  
\citet{zhang2019style} used a HIN to model marketplace vendor sybil\footnote{a single author can have multiple users accounts which are considered as \textit{sybils}} accounts on the darknet, where each node representing an object is associated with various features (e.g. content, photography style, user profile and drug information).
Similarly,~\citet{kumar2020edarkfind} proposed a multi-view unsupervised approach which incorporated features of text content, drug substances, and locations to generate vendor embeddings. We note that while such efforts~\cite{zhang2019style,kumar2020edarkfind} are related to our work, there are key distinctions. First, such efforts focus only on vendor sybil accounts. Second, in both cases, they rely on a host of multi-modal information sources (photographs, substance descriptions, listings, and location information) that are not readily available in our setting -  limited to forum posts.  Third, neither effort exploits multitask learning. 

\noindent {\bf Authorship Attribution of Short Text:}
\citet{kim-2014-convolutional} introduced convolutional neural networks (CNNs) for text classification.
Follow-up work on authorship attribution~\cite{ruder2016character, shrestha-etal-2017-convolutional} leveraged these ideas to demonstrate that CNNs outperformed other models, particularly for shorter texts. 
The models proposed in these works aimed at balancing the trade-off between vocabulary size and sequence length budgets based on tokenization at either the character or word level.
Further work on subword tokenization~\cite{sennrich-etal-2016-neural}, especially byte-level tokenization, have made it feasible to share vocabularies across data in multiple languages. 
Models built using subword tokenizers have achieved good performance on authorship attribution tasks for specific languages (e.g., Polish~\cite{grzybowski-etal-2019-sparse}) and also across multilingual social media data~\cite{andrews-bishop-2019-learning}.
Non-English as well as multilingual darknet markets have been increasing in number since 2013~\cite{ebrahimi2018detecting}.
Our work builds upon all these ideas by using CNN models and experimenting with both character and subword level tokens.

\noindent {\bf Multitask learning (MTL):}
MTL~\cite{caruana1997multitask}, aims to improve machine learning models' performance on the original task by jointly training related tasks. 
MTL enables deep neural network-based models to better generalize by sharing some of the hidden layers among the related tasks. 
Different approaches to MTL can be contrasted based on the sharing of parameters across tasks - strictly equal across tasks (hard sharing) or constrained to be close (soft-sharing)~\cite{Ruder2017AnOO}.
Such approaches have been applied to language modeling~\cite{howard-ruder-2018-universal}, machine translation~\cite{dong-etal-2015-multi}, and dialog understanding~\cite{rastogi-etal-2018-multi}.

%% file: sections/methodology.tex
\begin{figure}[!htbp]
    \centering
    \includegraphics[width=\linewidth]{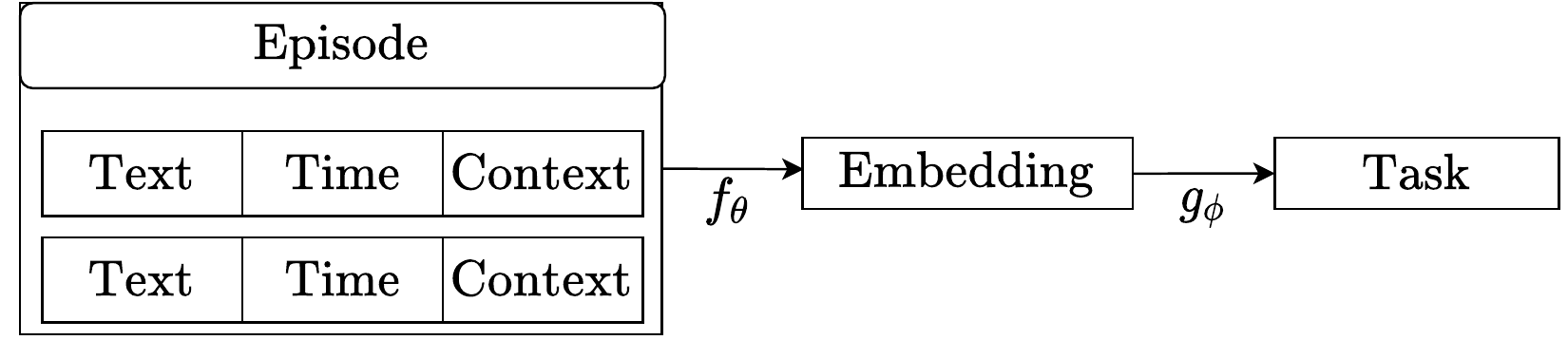}
    \caption{Overall \methodname{} Workflow.}
    \label{fig:main_workflow}
\end{figure}

Motivated by the success of social media user modeling using combinations of multiple posts by each user~\cite{andrews-bishop-2019-learning,Noorshams2020TIESTI}, we model posts on darknet forums using \textit{episodes}.
Each \textit{episode} consists of the textual content, time, and contextual information from multiple posts. 
A neural network architecture $f_{\theta}$ maps each episode to combined representation $e \in \mathbbm{R}^E$.
The model used to generate this representation is trained on various metric learning tasks characterized by a second set of parameters $g_{\phi}: \mathbbm{R}^E \xrightarrow[]{} \mathbbm{R}$.
We design the metric learning task to ensure that episodes having the same author have \textit{similar} embeddings.
Figure~\ref{fig:main_workflow} describes the architecture of this workflow and the following sections describe the individual components and corresponding tasks. 
Note that our base modeling framework is inspired by the social media user representations built by \citet{andrews-bishop-2019-learning} for a single task. 
We add meta-path embeddings and multitask objectives to enhance the capabilities of 
\methodname{}. 
Our implementation is available at: \url{https://github.com/pranavmaneriker/SYSML}.

\subsection{Component Embeddings}
Each episode $e$ of length $L$ consists of multiple tuples of texts, times, and contexts $e = \{(t_i, \tau_i, c_i) | 1 \leq i \leq L\}$. 
Component embeddings map individual components to vector spaces. 
All embeddings are generated from the forum data only; no pretrained embeddings are used.

\begin{figure}
    \centering
    \includegraphics[width=\linewidth]{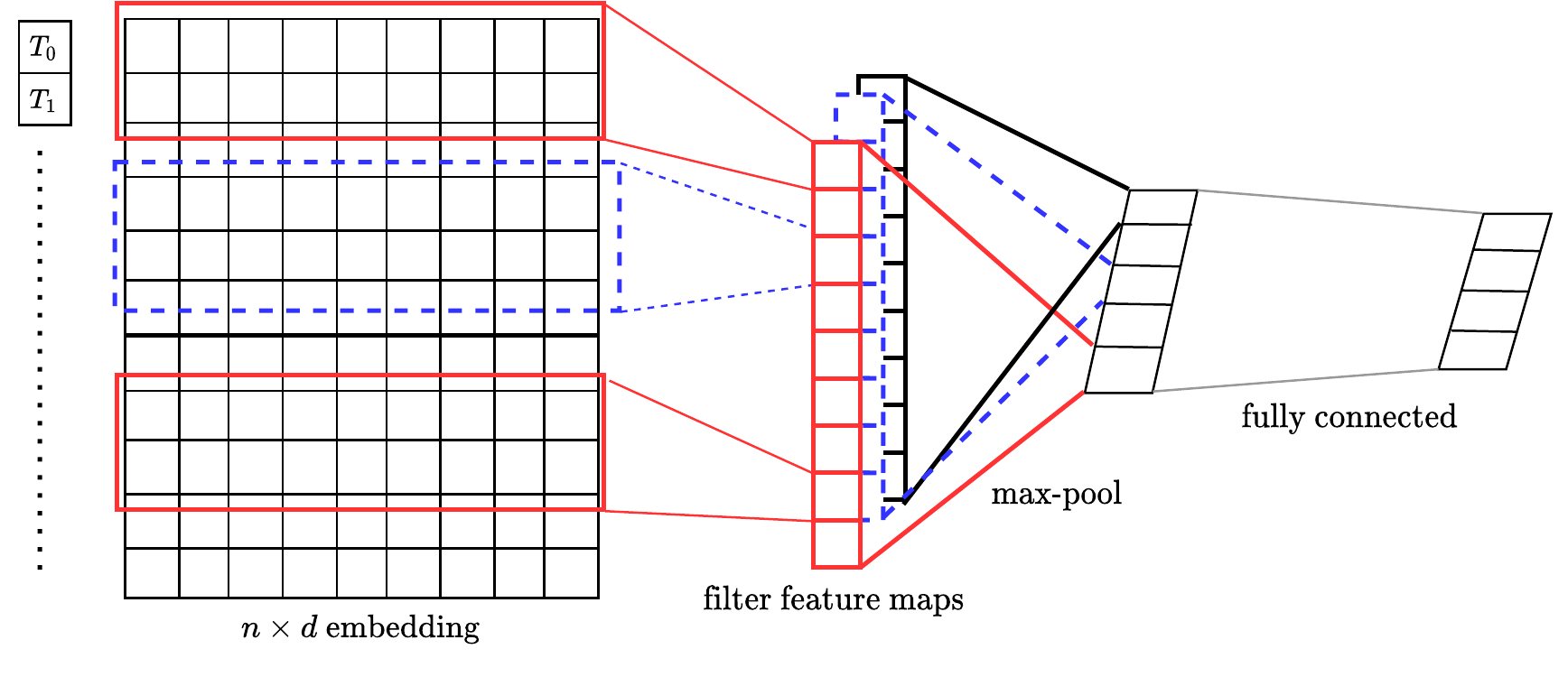}
    \caption{Text Embedding CNN \cite{kim-2014-convolutional}.}
    \label{fig:kim_cnn}
\end{figure}

\noindent \textbf{Text Embedding} First, we tokenize every input text post using either a character-level or byte-level tokenizer. 
A one-hot encoding layer followed by an embedding matrix $E_t$ of dimensions $|V| \times d_t$ where $V$ is the token vocabulary and $d_t$ is the token embedding dimension embeds an input sequence of tokens $T_0$, $T_1$, $\dots, T_{n-1}$.
We get a sequence embedding of dimension $n \times d_t$. 
Following this, we use $f$ sliding window filters, with filters sized $F = \{2, 3, 4, 5\}$ to generate feature-maps which are then fed to a max-over-time pooling layer, leading to a $|F| \times f$ dimensional output (one per filter).
Finally, a fully connected layer generates the embedding for the text sequence, with output dimension $d_t$. 
A dropout layer prior to the final fully connected layer prevents overfitting, as shown in Figure~\ref{fig:kim_cnn}. 

\noindent \textbf{Time Embedding} 
The time information for each post corresponds to when the post was created and is available at different granularities across darknet market forums. 
To have a consistent time embedding across different granularities, we only consider the least granular available date information (date) available on all markets.
We use the day of the week for each post to compute the time embedding 
by selecting the corresponding embedding vector of dimension $d_{\tau}$ from the matrix $E_w$.

\noindent \textbf{Structural Context Embedding} The context of a post refers to the threads that it may be associated with. 
Past work~\cite{andrews-bishop-2019-learning} used the subreddit as the context for a Reddit post.
In a similar fashion, we encode the subforum of a post as a one-hot vector and use it to generate a $d_c$ dimensional context embedding. 
In the previously mentioned work, this embedding is initialized randomly.
We deviate from this setup and use an alternative approach based on a \textit{heterogeneous graph} constructed from forum posts to initialize this embedding. 

\begin{definition}[Heterogeneous Graph]
A heterogeneous graph $G = (V, E, T)$ is one where each node $v$ and edge $e$ are associated with a `type' $T_i \in T$, where the association is given by mapping functions $\phi(v): V \rightarrow T_V$, $\psi(e): E \rightarrow T_E$, where $|T_V| + |T_E| > 2$
\end{definition}

\begin{figure}
    \centering
    \includegraphics[width=\linewidth]{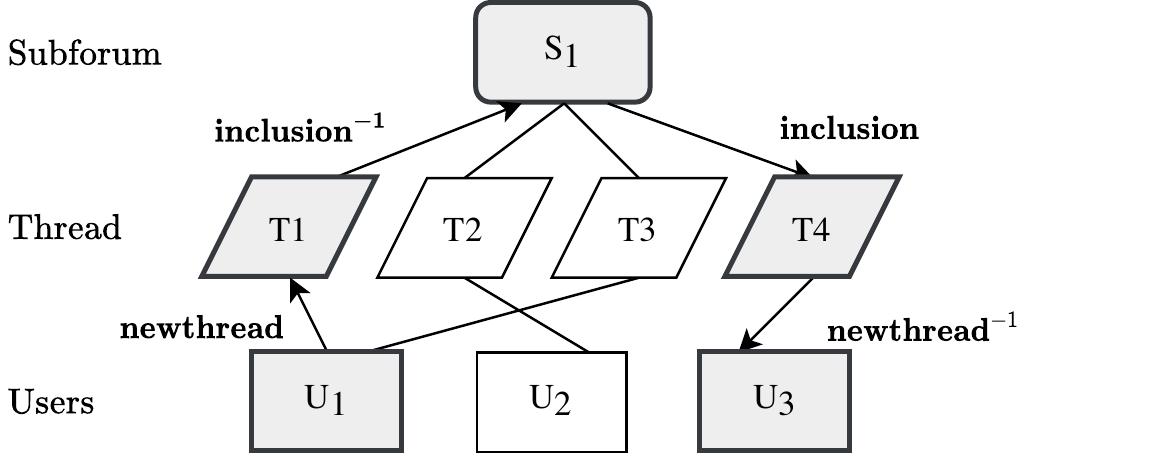}
    \caption{An instance of meta-path ‘UTSTU’ in a subgraph of the forum graph.}
    \label{fig:metapath}
\end{figure}

The constraint on $T_{V,E}$ ensures that at least one of $T_V$ and $T_E$ have more than one element (making the graph heterogeneous). Specifically, we build a graph in which there are four types of nodes: user (U), subforum (S), thread (T), and post (P), and each edge indicates either a post of new thread (U-T), reply to existing post (U-P) or an inclusion (T-P, S-T) relationship.
To learn the node embeddings in such heterogeneous graphs, we leverage the metapath2vec~\cite{dong2017metapath2vec} framework with specific meta-path schemes designed for darknet forums. 
Each meta-path scheme can incorporate specific semantic relationships into node embeddings. 
For example, Figure~\ref{fig:metapath} shows an instance of a meta-path ‘UTSTU’, which connects two users posting on threads in the same subforum and goes through the relevant threads and subforum.
Our analysis is user focused; to capture user behavior, we consider \emph{all} metapaths starting from and ending at a user node. Thus, to fully capture the semantic relationships in the heterogeneous graph, we use seven meta-path schemes: UPTSTPU, UTSTPU, UPTSTU, UTSTU, UPTPU, UPTU, and UTPU. As a result, the learned embeddings will preserve the semantic relationships between each subforum, included posts as well as relevant users (authors). 
Metapath2vec generates embeddings by maximizing the probability of heterogeneous neighbourhoods, normalizing it across typed contexts.
The optimization objective is:
\begin{align*}
            \arg\max\limits_\theta \prod\limits_{v\in V} \prod\limits_{t\in T_v} \prod\limits_{c_t \in N_t(v)} p(c_t|v; \theta)
\end{align*}
Where $\theta$ is the learned embedding, $N_t(v)$ denotes $v$'s neighborhood with the $t^{th}$ type of node. 
In practice, this is equivalent to running a word2vec~\cite{Mikolov2013EfficientEO} style skip gram model over the random walks generated from the meta-path schemes when $p(c_t|v; \theta) = $ is defined as a softmax function. Further details of metapath2vec can be found in the paper by \citet{dong2017metapath2vec}.

\subsection{Episode Embedding}
\begin{figure}
    \centering
    \includegraphics[width=\linewidth]{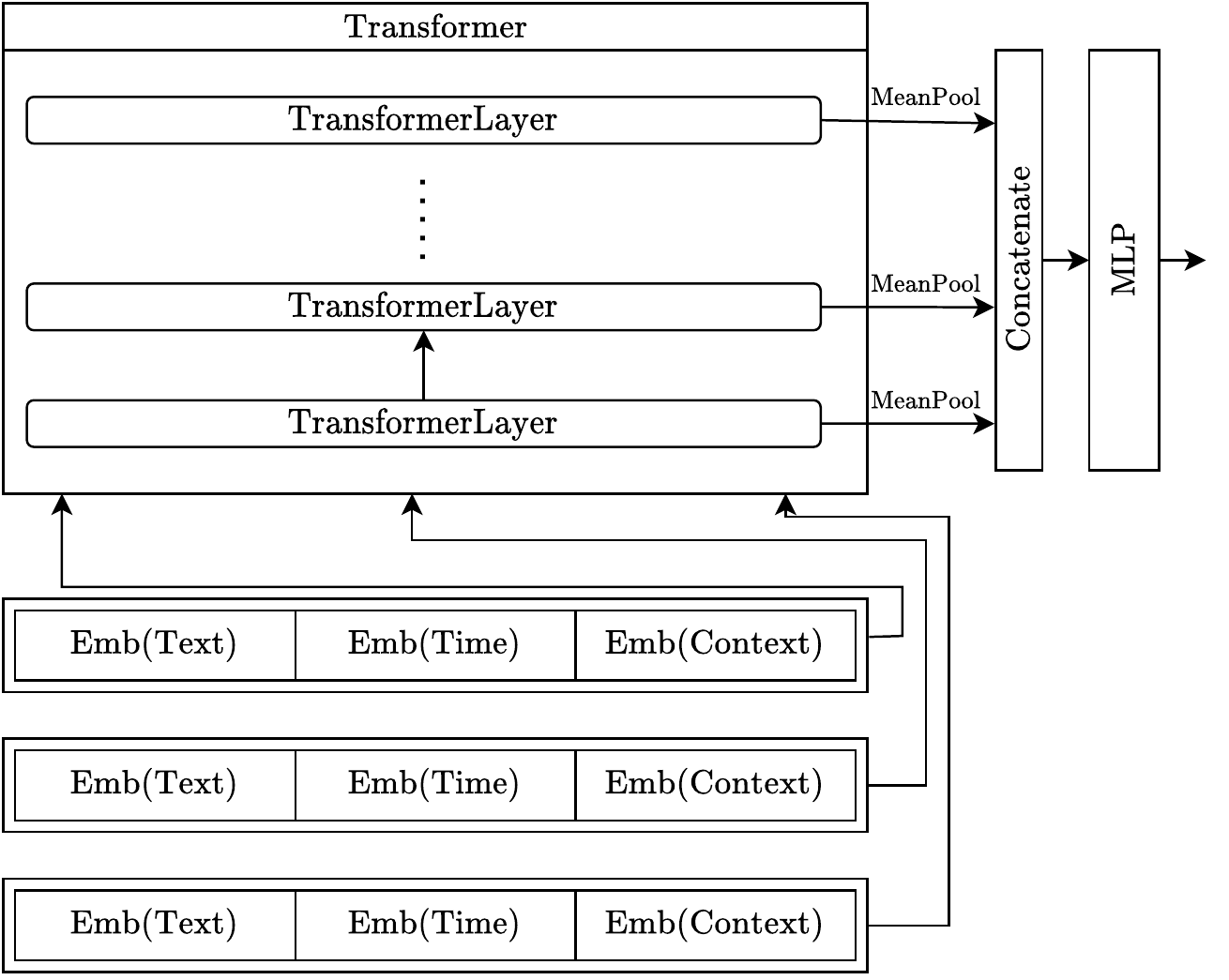}
    \caption{Architecture for Transformer Pooling.
    }
    \label{fig:emb_transformer}
\end{figure}

The embeddings of each component of a post are concatenated into a $d_e = d_t + d_{\tau} + d_c$ dimensional embedding.
An episode with $L$ posts, therefore, has a $L \times d_e$ embeddings. 
We generate a final embedding for each episode, given the post embeddings using two different models.
In \textbf{Mean Pooling}, the episode embedding is the mean of $L$ post embeddings, resulting in a $d_e$ dimensional episode embedding.
For the \textbf{Transformer}, the episode embeddings are fed as the inputs to a transformer model~\cite{devlin-etal-2019-bert,vaswani2017attention}, with each post embedding acting as one element in a sequence for a total sequence length $L$. 
We follow the architecture proposed by~\citet{andrews-bishop-2019-learning} and omit a detailed description of the transformer architecture for brevity (Figure~\ref{fig:emb_transformer} shows an overview).
Note that we do not use positional embeddings within this pooling architecture.
The parameters of the component-wise models and episode embedding models comprise the episode embedding $f_{\theta}: \left\{(t, \tau, c)\right\}^L \xrightarrow[]{} \mathbbm{R} ^ E$.

\subsection{Metric Learning}
\label{sec:framework:metric_learning}
An important element of our methodology is the ability to learn a distance function over user representations. We use the username as a label for the episode $e$ within the market $M$ and denote each username as a unique label $u \in U_M$.
Let $W =  |U_M| \times d_E $ represent a matrix denoting the weights corresponding to a specific metric learning method and let $x ^ * = \frac{x}{ || x||}$.
An example of a metric learning loss would be Softmax Margin, i.e., cross-entropy based softmax loss.
\begin{align*}
    P(u | e) = \frac{e^{W_{u} d_e}}{\sum\limits_{j=1}^{|U_M|}{e^{W_j d_e}}}
\end{align*}

We also explore alternative metric learning approaches such as Cosface (CF)~\cite{wang2018cosface}, ArcFace (AF)~\cite{deng2019arcface}, and MultiSimilarity (MS)~\cite{wang2019multi}. 

\subsection{Single-Task Learning}
The components discussed in the previous sections are combined together to generate  an embedding and the aforementioned tasks are used to train these models. 
Given an episode $e = \{(t_i, \tau_i, c_i) | 1 \leq i \leq L\}$, the componentwise embedding modules generate embedding for the text, time, and context, respectively.
The pooling module combines these embeddings into a single embedding $e \in \mathbbm{R}^E$. 
We define $f_\theta$ as the combination of the transformations that generate an embedding from an \textit{episode}.
Using a final metric learning loss corresponding to the task-specific $g_\phi$, we can train the parameters $\theta$ and $\phi$.
The framework, as defined in Figure~\ref{fig:main_workflow}, results in a model trainable for a single market $M_i$. 
Note that the first half of the framework (i.e., $f_\theta$) is sufficient to generate embeddings for episodes, making the module invariant to the choice of $g_\phi$. 
However, the embedding modules learned from these embeddings may not be compatible for comparisons across different markets, which motivates our multi-task setup. 

\subsection{Multi-Task Learning}

\begin{figure}
    \centering
    \includegraphics[width=\linewidth]{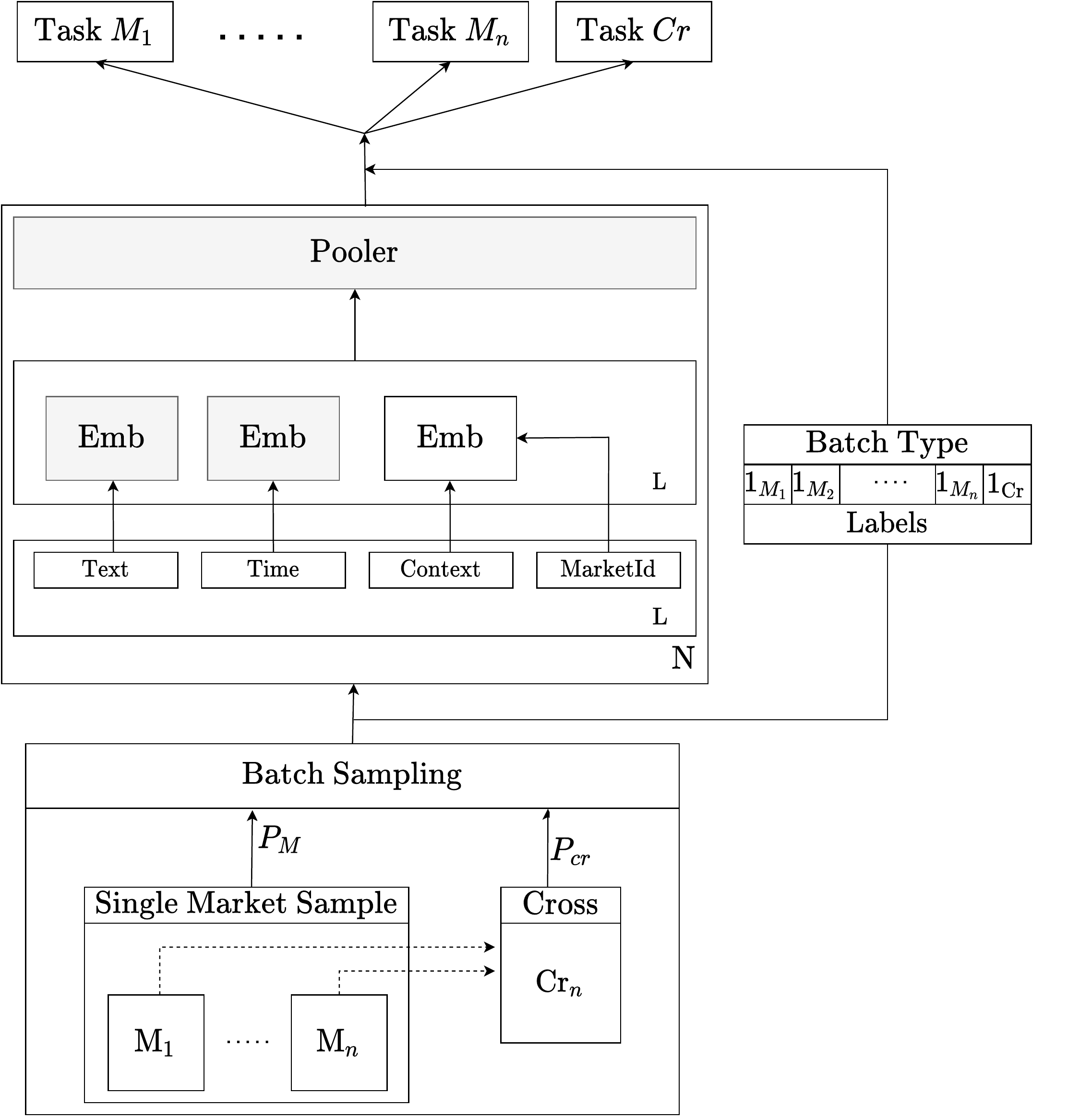}
    \caption{Multi-task setup. Shaded nodes are shared}
    \label{fig:multitask_setup}
\end{figure}

We use authorship attribution as the metric learning task for each market.
Further, a majority of the embedding modules are shared across the different markets.
Thus, in a multi-task setup, the model can share episode embedding weights (except context, which is market dependent) across markets. 
A shared BPE vocabulary allows weight sharing for text embedding on the different markets. 
However, the task-specific layers are not shared (different authors per dataset), and sharing $f_\theta$ does not guarantee alignment of embeddings across datasets (to reflect migrant authors). 
To remedy this, we construct a small, manually annotated set of labeled samples of authors known to have migrated from one market to another.
Additionally, we add pairs of authors known to be distinct across datasets.
The \textit{cross-dataset} consists of all episodes of authors that were manually annotated in this fashion.
The first step in the multi-task approach is to choose a market {($\mathcal{T}_M$)} or cross-market {($\mathcal{T}_{cr}$)} metric learning task $\mathcal{T}_i \sim \mathcal{T} =  \{\mathcal{T}_M, \mathcal{T}_{cr} \}$.
Following this, a batch of $N$ episodes $\mathcal{E} \sim \mathcal{T}_i$ is sampled from the corresponding task.
The embedding module generates the embedding for each episode $f_{\theta}^N: \mathcal{E} \xrightarrow[]{} \mathbbm{R}^{N \times E}$. 
Finally, the task-specific metric learning layer $g_{\phi}^{\mathcal{T}_i}$ is selected and a task-specific loss is backpropagated through the network. 
Note that in the \textit{cross-dataset}, new labels are defined based on whether different usernames correspond to the same author and episodes are sampled from the corresponding markets. 
Figure~\ref{fig:multitask_setup} demonstrates the shared layers and the use of \textit{cross-dataset} samples. 
The overall loss function is the sum of the losses across the markets: $\mathcal{L} = \underset{\mathcal{T}_i \sim \mathcal{T},\ \mathcal{E} \sim \mathcal{T}_i}{\mathbbm{E}}\left[\mathcal{L}_i(\mathcal{E})\right] $.

%% file: sections/dataset.tex
\label{sec:dataset}
\noindent \citet{munksgaard2016mixing} studied the politics of darknet markets using structured topic models on the forum posts across six  large markets. We start with this dataset and perform basic pre-processing to clean up the text for our purposes. We focus on four of the six markets - \textit{Silk Road} (\textbf{SR}), \textit{Silk Road 2.0}~({\textbf{SR2}}), \textit{Agora Marketplace}~(\textbf{Agora}), and \textit{Black Market Reloaded} (\textbf{BMR}). 
We exclude `The Hub'  as it is not a standard forum but an `omni-forum'~\cite{munksgaard2016mixing} for discussion of other marketplaces and has a significantly different structure, which is beyond the scope of this work.
We also exclude `Evolution Marketplace' since none of the posts had PGP information present in them and thus were unsuitable for migration analysis.

\noindent \textbf{Pre-processing} We add simple regex and rule based filters to replace quoted posts (i.e., posts that are begin replied to), PGP keys, PGP signatures, hashed messages, links, and images each with different special tokens (\texttt{[QUOTE]}, \texttt{[PGP PUBKEY]}, \texttt{[PGP SIGNATURE]}, \texttt{[PGP ENCMSG]}, \texttt{[LINK]}, \texttt{[IMAGE]}).
We retain the subset of users with sufficient posts to create at least two episodes worth of posts. In our analysis, we focus on episodes of up to 5 posts. 
To avoid leaking information across time, we split the dataset into approximately equal-sized train and test sets with a chronologically midway splitting point such that half the posts on the forum are before that time point.
Statistics for data after pre-processing is provided in Table~\ref{tab:dataset_stats}. Note that the test data can contain authors not seen during training.

\begin{table}[htbp]
\begin{scriptsize}
    \centering
    \begin{tabular}{ccccc}
    \toprule
         Market &  Train Posts & Test Posts & \#Users train & \#Users test \\
    \midrule
         SR & 379382 & 381959 & 6585 & 8865\\
         SR2 & 373905 & 380779 &  5346 & 6580 \\
         BMR & 30083 &30474 & 855 & 931\\
         Agora & 175978 & 179482 & 3115 & 4209\\
    \bottomrule
    \end{tabular}
    \caption{Dataset Statistics for Darkweb Markets.}
    \label{tab:dataset_stats}
    \end{scriptsize}
\end{table}

\noindent \textbf{Cross-dataset Samples} Past work has established PGP keys as strong indicators of shared authorship on darkweb markets~\cite{tai2019adversarial}. 
To identify different user accounts across markets that correspond to the same author, we follow a two-step process. 
First, we select the posts containing a PGP key, and then pair together users who have posts containing the same PGP key. 
Following this, we still have a large number of potentially incorrect matches (including scenarios such as information sharing posts by users sharing the PGP key of known vendors from a previous market). 
We manually check each pair to identify matches that clearly indicate whether the same author or different authors posted them, leading to approximately 100 reliable labels, with 33 pairs matched as migrants across markets.

%% file: sections/eval.tex
While ground truth labels for a single author having multiple accounts are unavailable, individual models can still be compared by measuring their performance on authorship attribution as a proxy. 
We evaluated our method using retrieval-based metrics over the embeddings generated by each approach. 
Denote the set of all episode embeddings as $E = \{e_1, \dots e_n\}$ and let $Q = \{q_1, q_2, \dots q_\kappa\} \subset E$ be the sampled subset.
We computed the cosine similarity of the query episode embeddings with all episodes. Let $R_i = \langle r_{i1}, r_{i2}, \dots r_{in} \rangle$ denote the list of episodes in $E$ ordered by their cosine similarity with episode $q_i$ (excluding itself) and let $A(.)$ map an episode to its author. The following measures are computed.

\noindent \textbf{Mean Reciprocal Rank}: (MRR) The RR for an episode is the reciprocal rank of the first element (by similarity) with the same author. MRR is the mean of reciprocal ranks for a sample of episodes.
\begin{align*}
    MRR(Q) = \frac{1}{\kappa}\sum_{i=1}^\kappa \frac{1}{\min\limits_j  \left(A(r_{ij}) = A(e_i)\right)}
\end{align*}

\noindent \textbf{Recall@k}:  (R@k)  Following~\citet{andrews-bishop-2019-learning}, we define the R@k for an episode $e_i$ to be an indicator denoting whether an episode by the same author occurs within the subset $ \langle r_{i1}, \dots, r_{ik} \rangle$. R@k denotes the mean of these recall values over all the query samples.

\noindent \textbf{Baselines}
We compare our best model against two baselines. First, we consider a popular short text authorship attribution model~\cite{shrestha-etal-2017-convolutional} based on embedding each post using character CNNs. 
While the method had no support for additional attributes (time, context) and only considers a single post at a time, we compare variants that incorporate these features as well. 
The second method for comparison is invariant representation of users~\cite{andrews-bishop-2019-learning}. This method considers only one dataset at a time and does not account for graph-based context information. 
Results for episodes of length 5 are shown in Table~\ref{tab:baselines_comparison}


\begin{table*}
    \small
    \centering
		\begin{tabular}{lcccccccc}
 		\toprule
			\multirow{2}{*}{Method}	&\multicolumn{2}{c}{BMR}	&	\multicolumn{2}{c}{Agora}	&	\multicolumn{2}{c}{SR2}	&	\multicolumn{2}{c}{SR}\\
					&MRR&	R@10&	MRR&	R@10&	MRR&	R@10&	MRR&	R@10\\
		\midrule
			\citet{shrestha-etal-2017-convolutional} (CNN) &	0.07	&	0.165	&	0.126	&	0.214	&	0.082	&	0.131	&	0.036	&	0.073	\\
			 + time + context &	0.235	&	0.413	&	0.152	&	0.263	&	0.118	&	0.21	&	0.094	&	0.178	\\
			 + time + context + transformer pooling &	0.219	&	0.409	&	0.146	&	0.266	&	0.117	&	0.207	&	0.113	&	0.205	\\
		\hline
			\citet{andrews-bishop-2019-learning} (IUR) \\
			 mean pooling &	0.223	&	0.408	&	0.114	&	0.218	&	0.126	&	0.223	&	0.109	&	0.19	\\
			 transformer pooling&	0.283	&	0.477	&	0.127	&	0.234	&	\textit{0.13}	&	\textit{0.229}	&	0.118	&	0.204	\\
		\hline
			\methodname{} (single) &	\textit{0.32}	&	\textit{0.533}	&	\textit{0.152}	&	\textit{0.279}	&	0.123	&	0.21	&	\textit{0.157}	&	\textit{0.266}	\\
			 - graph context &	0.265	&	0.454	&	0.144	&	0.251	&	0.089	&	0.15	&	0.049	&	0.094	\\
			 -graph context - time &	0.277	&	0.477	&	0.123	&	0.198	&	0.079	&	0.131	&	0.04	&	0.08	\\
		\hline
			\methodname{}  (multitask) &	\textbf{0.438}	&	\textbf{0.642}	&	\textbf{0.303}	&	\textbf{0.466}	&	\textbf{0.304}	&	\textbf{0.464}	&	\textbf{0.227}	&	\textbf{0.363}	\\
			 - graph context &	0.396	&	0.602	&	\textbf{0.308}	&	\textbf{0.469}	&	0.293	&	0.442	&	0.214	&	0.347	\\
			 - graph context - time &	0.366	&	0.575	&	0.251	&	0.364	&	0.236	&	0.358	&	0.167	&	0.28	\\
		\bottomrule
	\end{tabular}
    \caption{Best performing results in {\bf bold}. Best performing single-task results in {\it italics}. All  $\sigma_{MRR}< 0.02$, $\sigma_{R@10} < 0.03$, For all metrics, higher is better. Results suggest single-task performance largely outperforms the state-of-the-art \cite{shrestha-etal-2017-convolutional,andrews-bishop-2019-learning},
    while our novel multi-task cross-market setup offers a substantive lift ({\bf up to 2.5X on MRR and 2X on R@10}) over single-task performance.}
    \label{tab:baselines_comparison}
\end{table*}

%% file: sections/analysis.tex
\label{sec:analysis}

\subsection{Model and Task Variations}
To compare the variants using statistical tests, we compute the MRR of the data grouped by market, episode length, tokenizer, and a graph embedding indicator.
This leaves a small number of samples for paired comparison between groups, which precludes making normality assumptions for a t-test. 
Instead, we applied the paired two-samples Wilcoxon-Mann-Whitney (WMW) test~\cite{mann1947test}.
The first key contribution of our model is the use of meta-graph embeddings for context. 
The WMW test demonstrates that using pretrained graph embeddings was significantly better than using random embeddings ($p < 0.01$). 
Table~\ref{tab:baselines_comparison} shows a summary of these results using ablations.
For completeness of the analysis, we also compare the character and BPE tokenizers.
WMW failed to find any significant differences between the BPE and character models for embedding (table omitted for brevity). 
Many darkweb markets tend to have more than one language (e.g., BMR had a large German community), and BPE  allows a shared vocabulary to be used across multiple datasets with very few out-of-vocab tokens. 
Thus, we use BPE tokens for the forthcoming multitask models.
\begin{figure}[!htbp]
    \centering
    \includegraphics[width=0.8\linewidth]{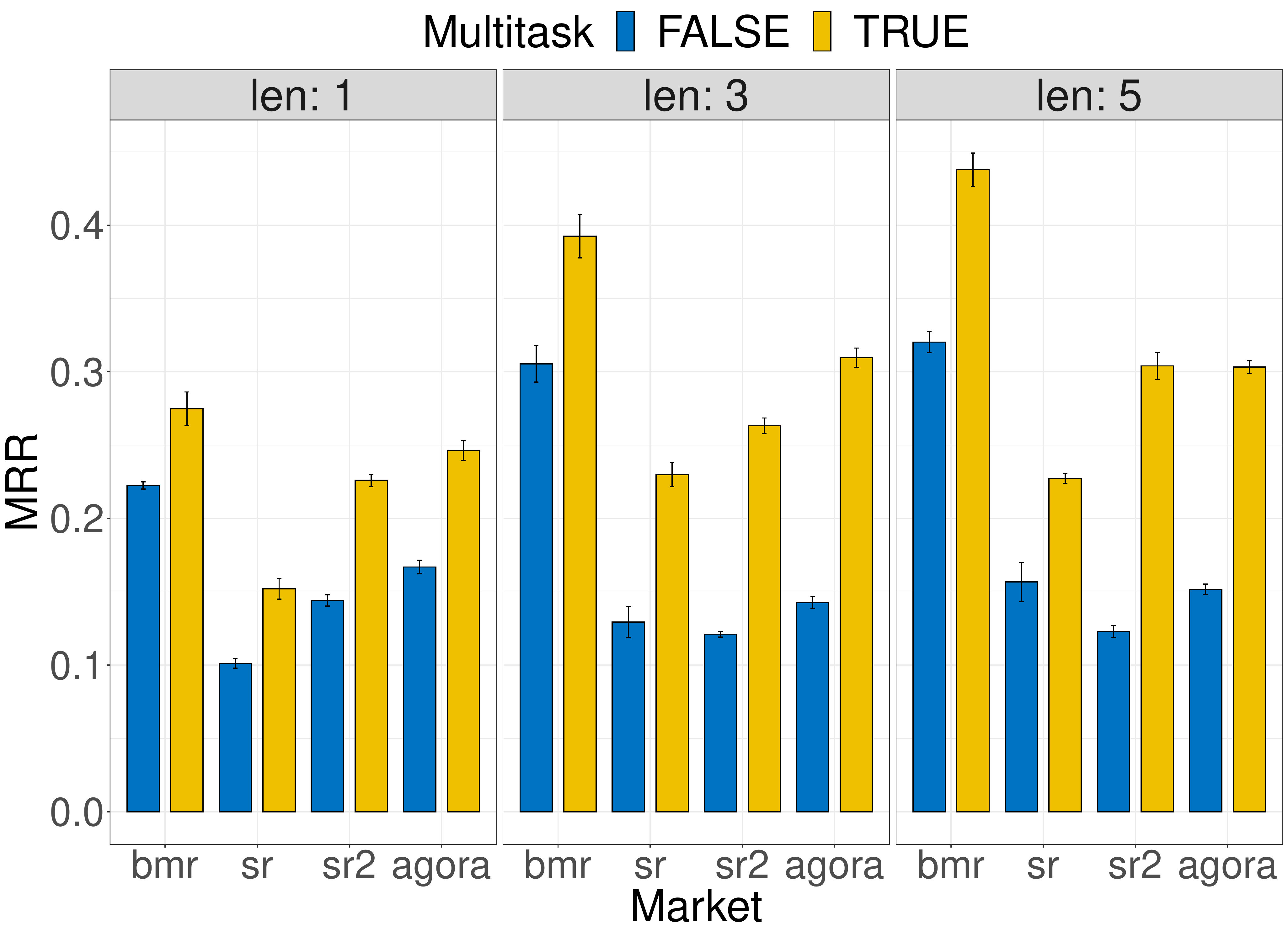}
    \caption{Drill-down: one-at-a-time vs. multitask.}
    \label{fig:multitask_results}
\end{figure} 

\noindent \textbf{Multitask} Our second key contribution is the multitask setup.
Table~\ref{tab:baselines_comparison} demonstrates that \methodname{} (multitask) outperforms all baselines on episodes of length 5. We further compare runs of the best single task model for each market against a multitask model. 
Figure~\ref{fig:multitask_results} demonstrates that multitask learning  consistently and significantly (WMW: $p < 0.01$) improves performance across all markets and all episode lengths.

\begin{figure}[!htbp]
    \centering
    \includegraphics[width=0.8\linewidth]{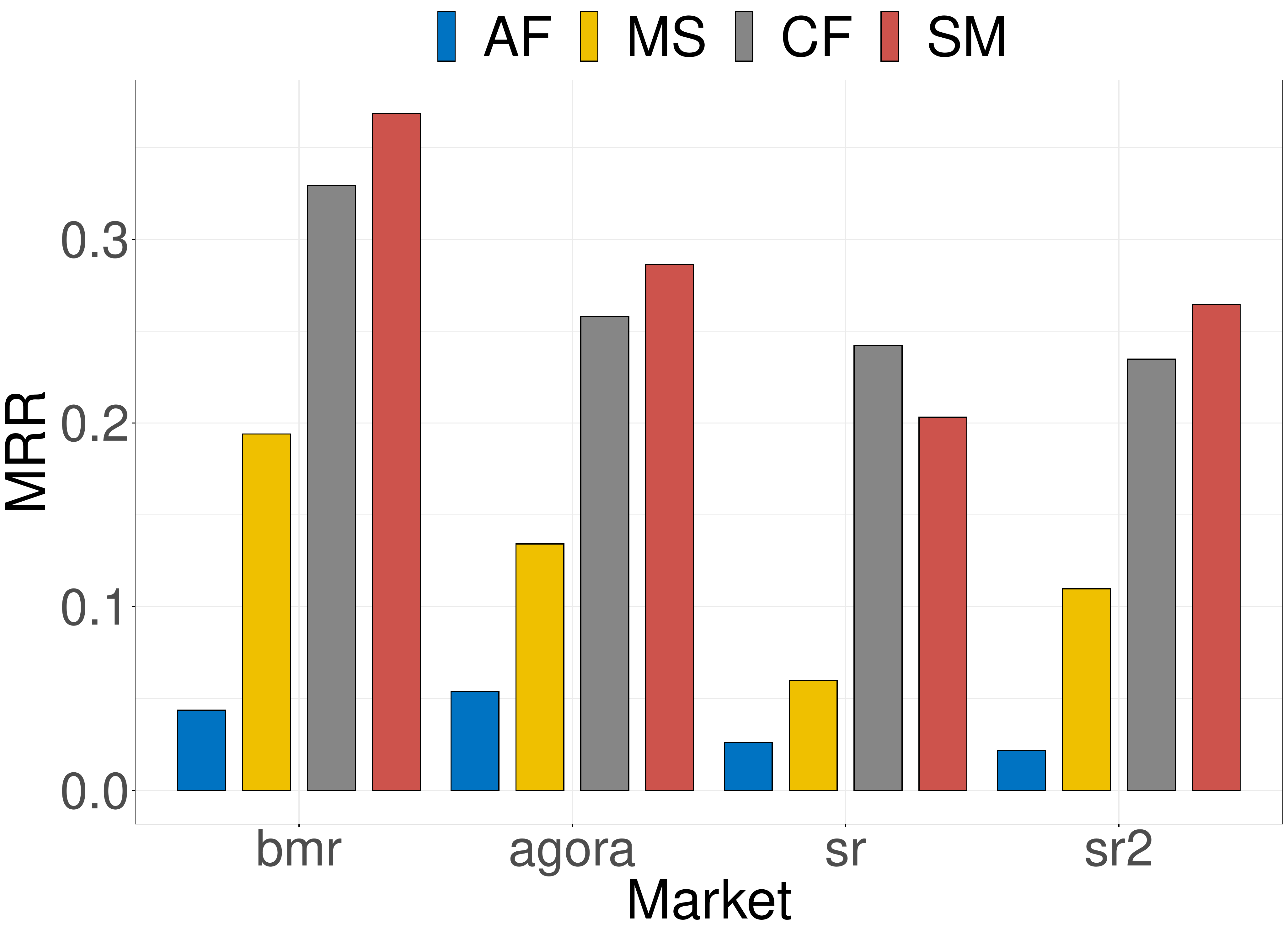}
    \caption{Task comparison: SM and CF are better performing two methods, with SM better in 3 of 4 cases.}
    \label{fig:metric_learning}
\end{figure}

\noindent \textbf{Metric Learning}
Recent benchmark evaluations have demonstrated that different metric learning methods provide only marginal improvements over classification~\cite{musgrave2020metric,Zhai2019ClassificationIA}. 
We experimented with various state-of-the-art metric learning methods (\S\ref{sec:framework:metric_learning}) in the multi task setup and found that softmax-based classification (SM) was the best performing method in 3 of 4 cases for episodes of length 5 (Figure~\ref{fig:metric_learning}). 
Across all lengths, SM is significantly better (WMW: $p < 1e-8$) and therefore we use SM in \methodname{}.

\subsection{Novel Users}
\begin{figure}
    \centering
    \includegraphics[width=0.8\linewidth]{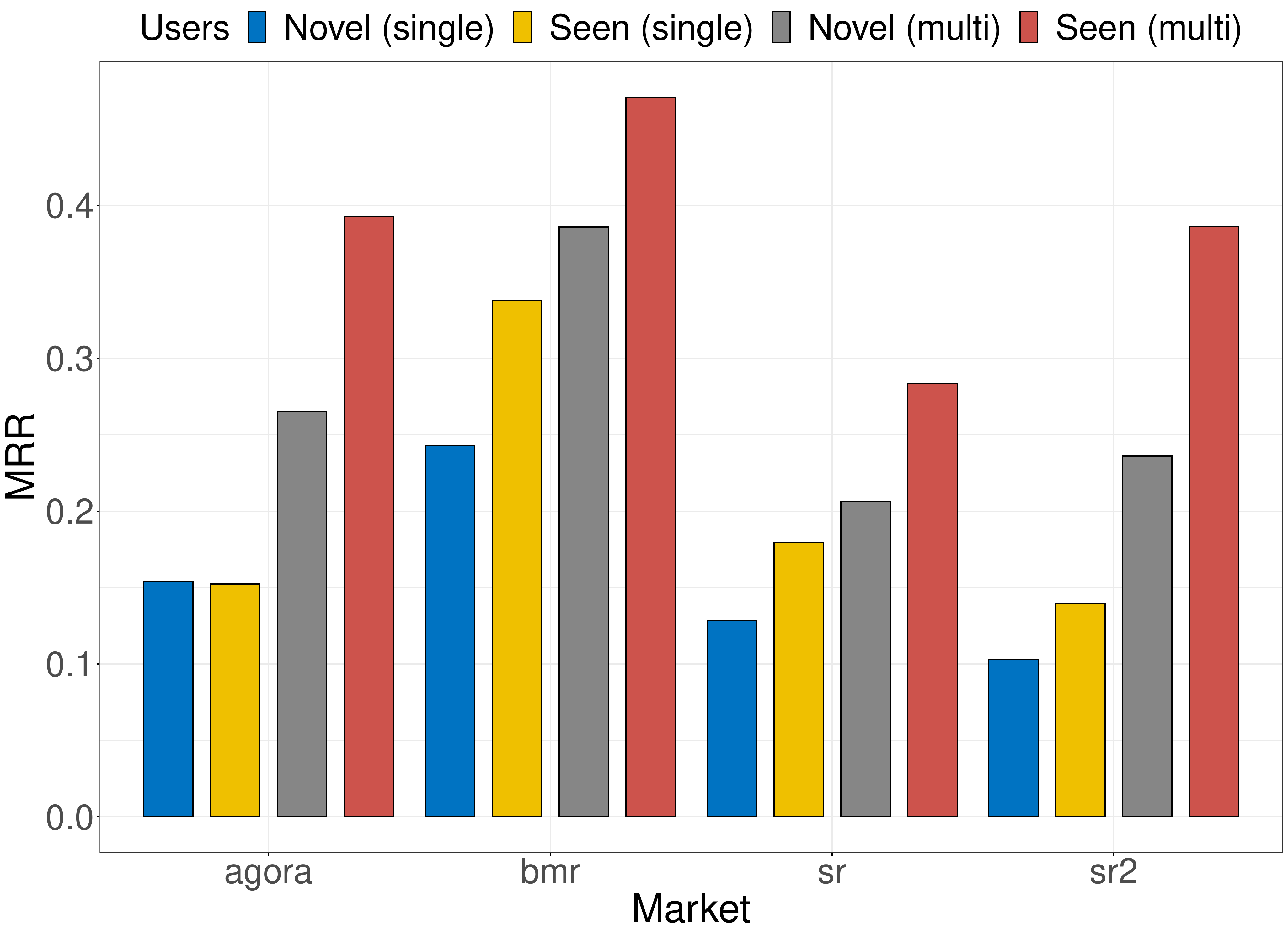}
    \caption{Lift on the multitask setup across users.}
    \label{fig:novel_vs_train_comparison}
\end{figure}

The dataset statistics (Table~\ref{tab:dataset_stats}) indicate that there are users in each dataset who have no posts in the time period corresponding to the training data. 
To understand the distribution of performance across these two configurations, we compute the test metrics over two samples.
For one sample, we constrain the sampled episodes to those by users who have at least one episode in the training period (Seen Users).
For the second sample, we sample episodes from the complement of the episodes that satisfy the previous constraint (Novel Users).
Figure~\ref{fig:novel_vs_train_comparison} shows the comparison of MRR on these two samples against the best single task model for episodes of length 5. 
Unsurprisingly, the first sample (Seen Users) have better query metrics than the second (Novel Users).
However, importantly both of these groups outperformed the best single task model results on the first group (Seen Users), which demonstrates that the {\it lift offered by the multitask setup is spread across all users.} 

\begin{figure}[!htbp]
    \centering
    \includegraphics[width=\linewidth]{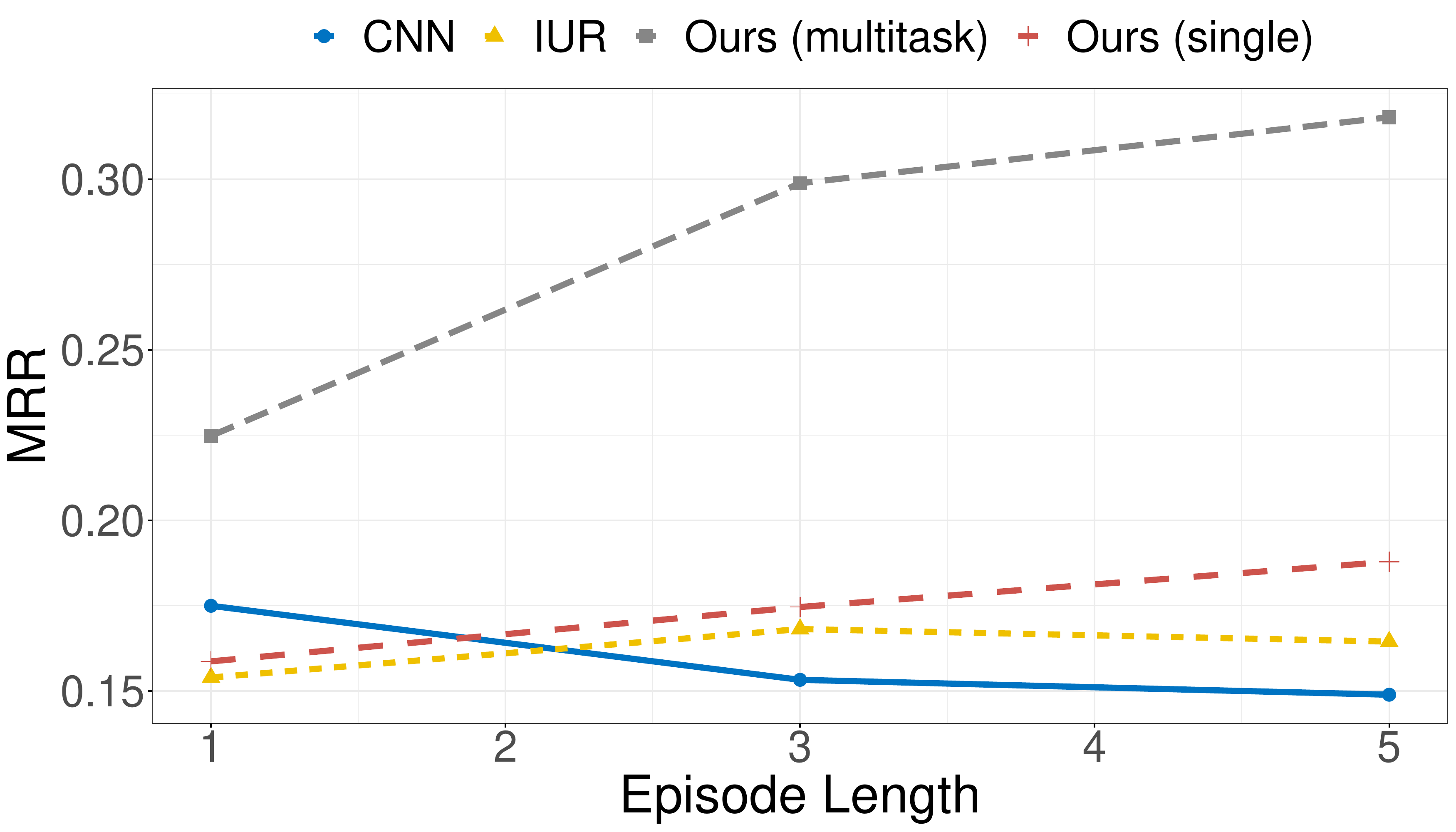}
    \caption{\methodname{} is more effective at utilizing multi post stylometric information}
    \label{fig:len_comparison}
\end{figure}

\noindent \textbf{Episode Length}  Figure~\ref{fig:len_comparison} shows a comparison of the mean performance of each model across various episode lengths. We see that compared to the baselines, \methodname{} can combine contextual and stylistic information across multiple posts more effectively. Additional results (see appendix),  indicate that this trend continues for larger episode sizes.

%% file: sections/casestudy.tex
\subsection{Qualitative Analysis of Attribution:}
\label{sec:casestudy:qualitative}
In this section, we consider the average (euclidean) distance between each pair of episodes by the same author as a heuristic for stylometric identifiability (SI), where lower average distance corresponds to higher SI and vice versa. 
Somewhat surprisingly, 
authors with a small number of total episodes ($<10$) were found at both extremes of identifiability, while the authors with the highest number of episodes were in the intermediate regions, suggesting that SI is not strongly correlated with episode length.  Next, we further investigate these groups.

\noindent \textbf{High SI authors:} 
Among the 20 users with the lowest average distance between episodes, a single pattern is prominent. 
This first group of high SI users are "newbie" users.  On a majority of analyzed forums, a minimum number of posts by a user is required before posting restrictions are removed from the user's account.
Thus, users create threads on `Newbie Discussion' subforums.
Typical posts on these threads include repeated posting of the same message or numbered posts counting up to the minimum required. 
As users tend to make all these posts within a fixed time frame, the combination of repeated, similar stylistic text and time makes the posts easy to identify. 
Exemplar episodes from this "newbie" group are shown in Table~\ref{tab:example_posts}. 

After filtering these users out, we identified a few more notable high SI users.
These include an author on BMR with frequent `\pounds' symbol and ellipses (`...') and an author on Agora who only posted referral links (with an eponymous username `ReferralLink'). 
Finally, restricting posts to those made by 200 most frequently posting users (henceforth, T200), we found a user (labeled HSI-Sec\footnote{pseudonym}) who frequently provided information on security, where character n-grams corresponding to `PGP', 'Key', 'security' are frequent (Table~\ref{tab:attribution}). 
Thus, \methodname{} is able to leverage vocabulary and punctuation-based cues for SI.

\noindent \textbf{Low SI authors:}  Here, we attempt to characterize the post episode styles that are challenging for \methodname{} to attribute to the correct author.
Seminal work by~\citet{brennan2009practical, brennan2012adversarial} has demonstrated that obfuscation and imitation based strategies are effective against text stylometry.
We analyze the T200 authors who had high inter-episode distances to ascertain whether this holds true for \methodname{}.
For the least (and third least) identifiable author among T200, we find that frequent word n-grams are significantly less frequent than those for the most identifiable author from this subset (most frequent token occurs $\sim 600$ times vs. $\sim 4800$ times for identifiable) despite having more episodes overall. 
Further, one of the most frequent tokens is the \texttt{{[QUOTE]}} token, implying that this author frequently incorporates other authors' quotes into their posts.  
This strategy is analogous to the imitation based attack strategy proposed by~\citet{brennan2012adversarial}.
For the second least identifiable T200 author, we find that the frequent tokens have even fewer occurrences, and the special token \texttt{{[IMAGE]}} and its alternatives are among the frequent tokens - suggesting that an obfuscation strategy based on diversifying the vocabulary is effective.
Some samples are presented in Table~\ref{tab:attribution} under LSI-1 and LSI-2.

\begin{table}[]
    \tiny
    \centering
    \begin{tabularx}{\linewidth}{XX}
    \toprule
        Thread & Posts  \\
    \midrule
         Spam to 50 \&  Get out of Noobville &  26, 27, 28, 29, 30 \\
         Post 30 Times $\dots$ To Post Anywhere& 7, 8, 9, $\dots$\\ 
         Spam to 50 $\dots$ & 46, 47, $\dots$, Yeah 50 Spam! \\
         $\dots$ use my link $\dots$ & {[LINK]}, Here is my ref link {[LINK]}, Try this link {[LINK]}, $\dots$\\
    \bottomrule
    \end{tabularx}
    \caption{Examples of highly identifiable posts.}
    \label{tab:example_posts}
\end{table}
\input{sections/attr_table}
\noindent \textbf{Gradient-based attribution:} To cement our preceding hypotheses, 
we investigate whether the generated embedding can be attributed to phrases in the input which were mentioned in the previous section. 
We use Integrated Gradients~\cite{sundararajan2017axiomatic}, an axiomatic approach to input attribution.
Integrated Gradients assign an importance score to each feature which corresponds to an approximation of the integral of the gradient of a model's output with respect to the input features along a path from some reference baseline value (in our case, all \texttt{[PAD]} tokens) to the  input feature.
In Table~\ref{tab:attribution}, the highlight color corresponds to the attribution importance score for the presented posts.
We observed that the attribution scores correspond to our intuitions: HSI-Sec had high importance for security words, LSI-1 had obfuscated posts due to the presence of common image tokens, and LSI-2 had quotes mixed in, lead to misattribution (imitation-like strategy).

\begin{figure}
    \centering
    \includegraphics[width=\linewidth]{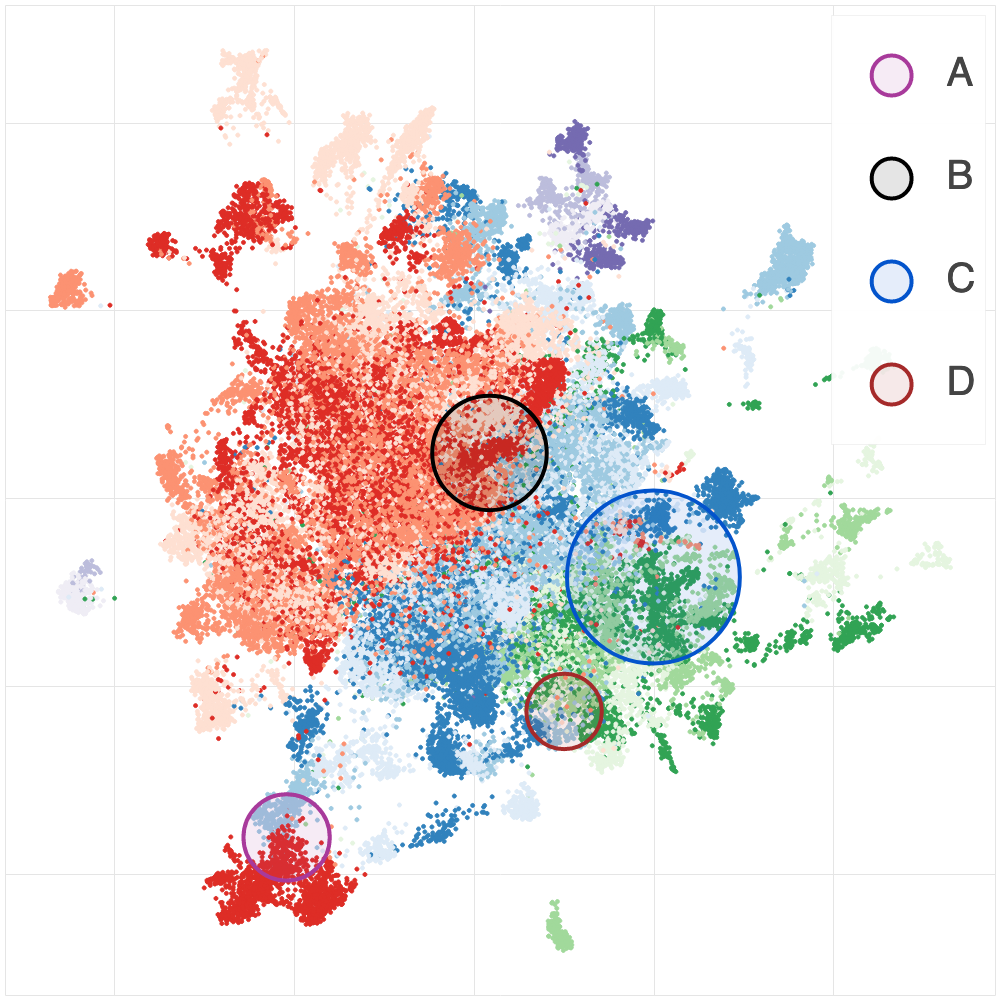}
    \caption{UMAP visualization of cross dataset embeddings for the top 200 authors, one hue per market. Circles denote the same user in two different markets. 
    }
    \label{fig:cross_dataset}
\end{figure}

\subsection{Migrant Analysis}
To understand the quality of alignment from the episode embeddings generated by our method, we use a simple top-k heuristic: for each episode of a user, find the top-k nearest neighboring episodes from other markets, and count the most frequently occurring user among these (candidate sybil account).
Figure~\ref{fig:cross_dataset} shows a UMAP projection for T200. 
Users of each market are colored by sequential values of a single hue (i.e., reds - SR2, blues - SR, etc.).
The circles in the figure highlight the top four pairs of users (top candidate sybils) with a frequent near neighbor from a different market. 
We find that each of these pairs can be verified as sybil accounts, either by a shared username (A, C, D) or by manual inspection of posted information (B).   
Note that none of these pairs were pre-matched using PGP - none were present in the high-precision matches. Thus,
\methodname{} is able to identify high ranking sybil matches reflecting users that migrate from one market to another.

%% file: sections/attr_table.tex
\newcolumntype{s}{>{\hsize=.1\hsize}X}
\begin{table}
\tiny
\begin{tabularx}{\linewidth}{sX}
\toprule
Author & Word Importance \\
\midrule
 & $\dots$ 2 cents, anyway $\dots$ 
PGP Key\colorbox[rgb]{0.8424999999999999, 0.9775000000000001, 0.8424999999999999}{ Fingerprint} = $\dots$\\
HSI-Sec & $\dots$ 
PGP Key\colorbox[rgb]{0.9124999999999999, 0.9875, 0.9124999999999999}{ Fingerprint} $\dots$ \colorbox[rgb]{0.8600000000000002, 0.9799999999999999, 0.8600000000000002}{...} $\dots$ security is NOT\colorbox[rgb]{0.8074999999999999, 0.9725000000000001, 0.8074999999999999}{ retroactive}. 
\\
& $\dots$ Is it possible for a\colorbox[rgb]{0.8775, 0.9825000000000002, 0.8775}{ gpg}\colorbox[rgb]{0.9475, 0.9924999999999999, 0.9475}{ key}\colorbox[rgb]{0.9924999999999999, 0.9475, 0.9475}{ to} request that ) \\
\hline
 & Check out the\colorbox[rgb]{0.985, 0.8949999999999999, 0.8949999999999999}{ link} in my sig $\dots$ \colorbox[rgb]{0.9475, 0.9924999999999999, 0.9475}{ [}\colorbox[rgb]{0.9475, 0.9924999999999999, 0.9475}{IMAGE} alt=8)]\\
LSI-1 & Hey dude, just run a search $\dots$\colorbox[rgb]{0.9875, 0.9124999999999999, 0.9124999999999999}{ I} can not help much $\dots$ \colorbox[rgb]{0.9875, 0.9124999999999999, 0.9124999999999999}{ Im} sure if you ask $\dots$ German\colorbox[rgb]{0.9475, 0.9924999999999999, 0.9475}{,} he may be willing to lend a hand. Good luck freind\colorbox[rgb]{0.7024999999999998, 0.9575000000000001, 0.7024999999999998}{ [}\colorbox[rgb]{0.4049999999999999, 0.9150000000000001, 0.4049999999999999}{IMAGE} alt=8)]\\
\midrule 
& [\colorbox[rgb]{0.9924999999999999, 0.9475, 0.9475}{QUOTE}] From: $\dots$ Just my opinion,\colorbox[rgb]{0.9124999999999999, 0.9875, 0.9124999999999999}{ I}\colorbox[rgb]{0.8949999999999999, 0.985, 0.8949999999999999}{'ve} done just about everything, $\dots$ \colorbox[rgb]{0.7900000000000001, 0.9699999999999999, 0.7900000000000001}{IMAGE} alt=8)] couldnt agree more\\
LSI-2 & [\colorbox[rgb]{0.9299999999999999, 0.99, 0.9299999999999999}{QUOTE}]\colorbox[rgb]{0.8250000000000002, 0.9749999999999999, 0.8250000000000002}{ From}\colorbox[rgb]{0.8775, 0.9825000000000002, 0.8775}{:} $\dots$  \colorbox[rgb]{0.9475, 0.9924999999999999, 0.9475}{ strangely} enough, when im in $\dots$ I too jabber\colorbox[rgb]{0.985, 0.8949999999999999, 0.8949999999999999}{ meaningless} jibberish $\dots$ \\
\bottomrule 
&Negative \fcolorbox{black}[rgb]{0.9000000000000001, 0.2999999999999998, 0.2999999999999998}{\rule{0pt}{2pt}\rule{2pt}{0pt}} Neutral \fcolorbox{black}[rgb]{1.0, 1.0, 1.0}{\rule{0pt}{2pt}\rule{2pt}{0pt}} Positive \fcolorbox{black}[rgb]{0.125, 0.875, 0.125}{\rule{0pt}{2pt}\rule{2pt}{0pt}}
\end{tabularx}
\caption{Integrated Gradient based attribution of posts}
\label{tab:attribution}
\end{table}


%% file: sections/conclusion.tex
We develop a novel stylometry-based multitask learning approach that leverages graph context to construct low-dimensional representations of short episodes of user activity for authorship and identity attribution. Our results on four different darknet forums suggest that both graph context and multitask learning provides a significant lift over the state-of-the-art. 
In the future, we hope to evaluate how such methods can be levered to analyze how users maintain trust while retaining anonymous identities as they migrate across markets. 
Further, we hope to quantitatively evaluate the migration detection to assess the evolution of textual style and language use on darknet markets. 
Characterizing users from a small number of episodes has important applications in limiting the propagation of bot and troll accounts, which will be another direction of future work.

%% file: sections/ack.tex

This work has been supported by NSF grants SES-1949037, CCF-2028944, and OAC-2018627.
All content represents the opinion of the authors, and is not necessarily endorsed by their sponsors.
The authors thank the OSU CLIPPERS group, Micha Elsner, Sean Current, Saket Gurukar, and anonymous reviewers for helpful discussions and feedback on this work. 
Additionally, the authors thank the Ohio Supercomputing Center~\cite{OhioSupercomputerCenter1987} and the OSU RI2 cluster for providing the computational resources used for conducting the experiments in this paper.


%% file: sections/ethics.tex
\label{sec:ethics}
The research conducted in this study was deemed to be {\it exempt research} by the Ohio State University's Office of Responsible Research Practices, since the forum data is classified as 'publicly available'. 
Darknet forum data is readily available publicly across multiple markets~\cite{dnmArchives,munksgaard2016mixing} and we follow standard practices for the darkweb~\cite{kumar2020edarkfind} limiting our analysis to publicly available information only. 
The data was originally collected to study the prevalence of illicit drug trade and the politics surrounding such trades.

\noindent \textbf{Limiting Harm} To the best of our knowledge, the collected data does not contain leaked private information~\cite{munksgaard2016mixing}. Beyond relying on the exempt nature of the study, we also strive to take further steps for minimizing harms from our research.
In accordance with the ACM Code of Ethics and to limit potential harm, we carry out substantial pre-processing (\S\ref{sec:dataset}) to remove links, images, and keys that may contain sensitive information.  
Towards respecting the privacy of subjects, we do not connect the identity of users to any private information; our method serves only to link users across markets.
Further, in this study, we restrict our analysis to darknet markets that have been inactive for several years. 
The darknet market community has itself taken steps over the past few years to link identities of trustworthy members across market closure via development of information hubs such as Grams, Kilos, and Recon~\cite{broadhurst2021impact}. 
Our efforts aim to understand the formative years that lead towards this centralization.

\noindent \textbf{Inclusiveness} Our methods do not attempt to characterize any traits of the users making the posts. Based on our analysis, the datasets contain posts in English, German, and Italian. 
Thus, our methods may be limited in applicability and biased in performance for languages belonging to these and related Indo-European languages.  

\noindent \textbf{Potential for Dual Use} Our goal is to understand how textual style evolves on darknet markets and how users on such markets may misuse them for scams and illicit activities. This digital forensic analysis can be put to good use for understanding trust signalling on these markets.
We understand the potential harm from dual use; stylometric methods could be used for the identification of users who may not want their identity to be made public, especially when they are subject of hostile governments.
We believe that making the information about the existence of such stylometric advances public and providing prescriptions for avoidance techniques (\S\ref{sec:casestudy:qualitative}) would aid users who may not know of strategies that they can use to preserve their anonymity.
Existing work~\cite{Noorshams2020TIESTI,andrews-bishop-2019-learning} has already expanded the use of stylometry to the open web. 
Thus, we have made the analysis of patterns that lower stylometric identifiability one focus of our case study.


%% file: sections/supplementary/supp.tex
\section{Reproducibility}
We describe the various hyperparameter settings used for the models trained by us. All deep learning models are implemented in python using Pytorch\footnote{https://pytorch.org/}, and the original C++ implementation of metapath2vec is used for generating metapath embeddings\footnote{https://ericdongyx.github.io/metapath2vec/m2v.html}. We used an implementation from the Captum python library~\cite{kokhlikyan2020captum} that uses the Gauss-Legendre quadrature rule for approximating the gradient.

\section{Training Hyperparameters}
We use batches of size 256. The Adam optimizer is used for training each network. The initial learning rate is set to $1e-3$, with a multiplicative decay factor of $0.5$ if the validation metrics do not improve after 5 epochs. Each model is trained for 30 epochs, and each configuration is run 5 times. 
We used a V100 GPU to train each run, with the average running time of 27:17 per run (mm:ss). For each run, 10\% of the dataset is used for validation. The best model is selected on the basis of minimum validation loss.

\subsection{Model Hyperparameters}

\subsubsection{Text Embedding Model}
Character vocabularies of size 1k and BPE vocabularies of size 30k are trained using only training portion of the datasets. The HuggingFace Tokenizers\footnote{https://github.com/huggingface/tokenizers} library is used to build the byte-level BPE vocab.
We use a Text CNN for embedding text across all settings. Each token has $32$ dimensional embeddings, and the final embedding dimension for a text sequence is set to $128$. Filters of sizes $\{2, 3, 4, 5\}$ are used, and the dropout probability is set to $0.1$ for the final layer.
\subsubsection{Time Embedding}
The time embedding dimension is set to 64.

\subsubsection{Context Embedding}
The context embedding dimension is set to $128$. For metapath2vec, we generate $1000$ walks for each user (author) node, the number of negative samples for each user is $5$, and the window size in the skip-gram model is set to $7$. These hyperparameters are also used in the metapath2vec work. For the length of each sampled walk, we set it to $80$, which is widely used in many representative skip-gram based embedding methods such as node2vec.

\subsubsection{Pooling Transformer}
The pooling transformer model has a feed forward layer dimension and final dimension of 128. There are 4 layers, each with 4 heads. The droupout probability for the final feed forward leyer layer is set to 0.1, and the output dimension is set to 32.

\subsubsection{Metric Learning Techniques}
We use the pytorch metric learning\footnote{https://ericdongyx.github.io/metapath2vec/m2v.html} package for implementing the different metric learning approaches, with the default parameters for each approach from their corresponding papers. i.e.,
\begin{itemize}
    \item \textbf{CosFace}: $m = 0.35$, $s = 64$
    \item \textbf{ArcFace}: $m = 28.6^{\circ}$, $s = 64$
    \item \textbf{MultiSimilarity}: $\alpha = 2$, $\beta = 50$, $\lambda = 0.5$
\end{itemize}

\section{Parameter Search}
Most hyperparameter comparisons are reported in the paper. For the multitask dataset sampling,
we ran the multitask model with $P_{cr} \in \{0.01, 0.02, 0.04, 0.1\}$, with $P_M = 1 - P_{cr}$, $P_{M_I} \propto |M_i|$ with similar performances up to $P_{cr} = 0.04$ and a drop at $P_{cr} = 0.1$. 
All results reported in the paper have $P_{cr} = 0.01$

\section{Metrics}
All metrics are computed using a sample of the episode embeddings. The sample size used for computing the metrics is $\kappa=1000$

\section{Additional Results}
\begin{figure}
    \centering
    \includegraphics[width=\linewidth]{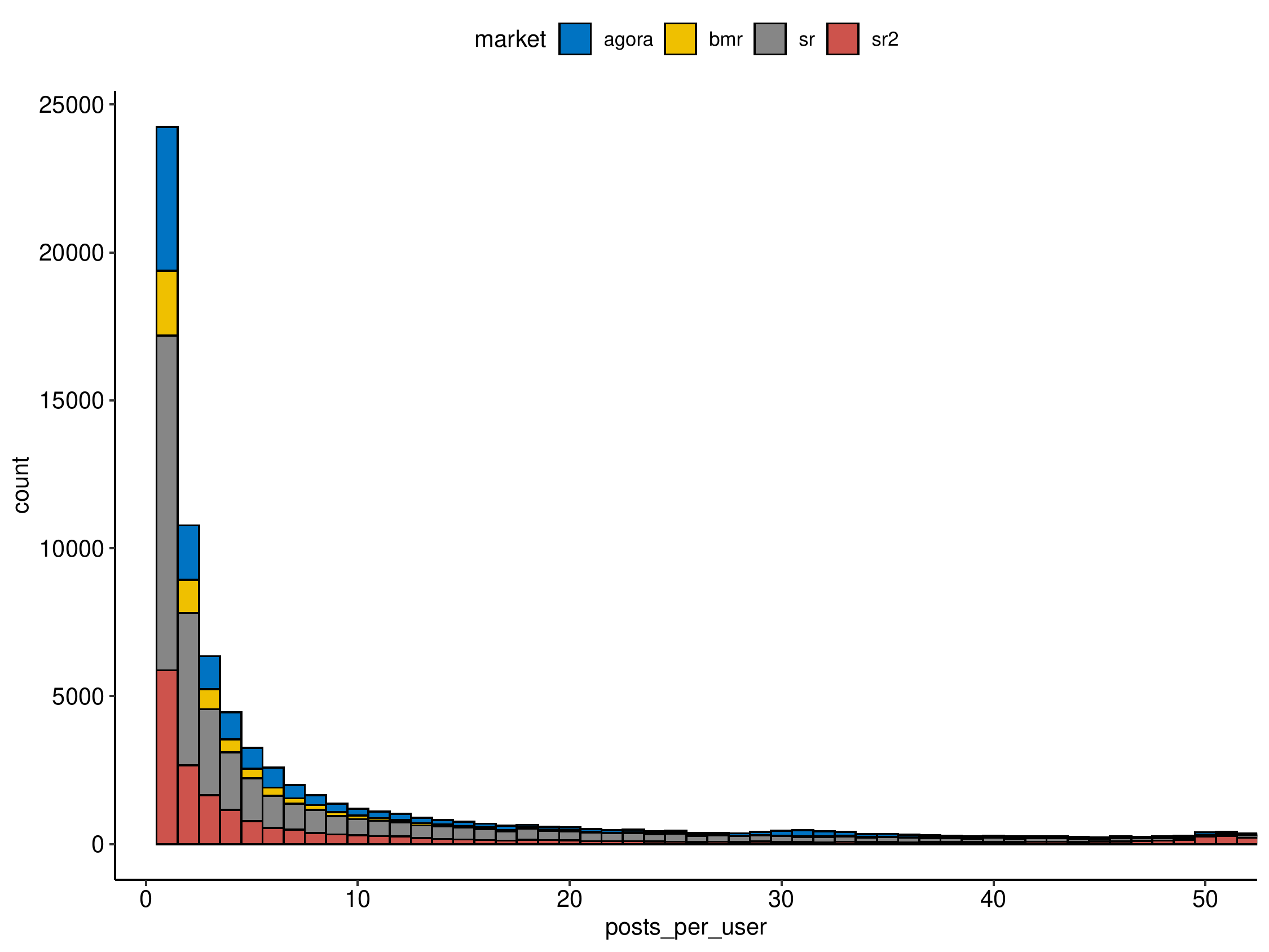}
    \caption{Frequency of number of posts per user}
    \label{fig:posts_freq}
\end{figure}

\begin{table*}[!htbp]
    \small
    \centering
		\begin{tabular}{lcccccccc}
 		\toprule
			\multirow{2}{*}{Method}	&\multicolumn{2}{c}{BMR}	&	\multicolumn{2}{c}{Agora}	&	\multicolumn{2}{c}{SR2}	&	\multicolumn{2}{c}{SR}\\
					&MRR&	R@10&	MRR&	R@10&	MRR&	R@10&	MRR&	R@10\\
		\midrule
			\methodname{}  (singletask) &	0.305	&	0.508	&	0.186	&	0.32	&	0.159	&	0.273	&	0.14	&	0.246	\\
				\hline
			\methodname{}  (multitask) &	0.484	& 0.689	&	0.349	&	0.519	&	0.401	&	0.556	&	0.292	&	0.429	\\
					\bottomrule
    	\end{tabular}
    	\caption{Additional results for 7 posts per episode}
      \label{tab:additional_res_7}
\end{table*}

\begin{table*}[!htbp]
    \small
    \centering
		\begin{tabular}{lcccccccc}
 		\toprule
			\multirow{2}{*}{Method}	&\multicolumn{2}{c}{BMR}	&	\multicolumn{2}{c}{Agora}	&	\multicolumn{2}{c}{SR2}	&	\multicolumn{2}{c}{SR}\\
					&MRR&	R@10&	MRR&	R@10&	MRR&	R@10&	MRR&	R@10\\
		\midrule
			\methodname{}  (singletask) &	0.264 & 0.48	&	0.146	&	0.249	&	0.165   &	0.272	&	0.194	&	0.319	\\
				\hline
			\methodname{}  (multitask) &	0.4667	&	 0.648	&	0.357	&	0.498	&	0.377	&	0.522	&	0.299	&	0.449	\\
					\bottomrule
    	\end{tabular}
    	\caption{Additional results for 9 posts per episode}
      \label{tab:additional_res_9}
\end{table*}

From Figure~\ref{fig:posts_freq}, we see that the number of users reduces rapidly as the posts per user decrease. Thus, we limited our analysis to up to 5 posts per episode.
For completeness, we also provide additional results for 7 and 9 posts per episode in Table~\ref{tab:additional_res_7} and ~\ref{tab:additional_res_9} respectively. Note that the histogram has some non-smooth bumps at around 10, 50, 100 posts as they act as the minimum number of posts for different levels of forum users. As explained in a previous section, users post on `newbie' forums until they reach a specific number of posts, leading to these unusual bumps in the histogram. 
We note that the performance of our methods continues to improve as the posts per episode are increased (at a cost to coverage - number of users studied), though the improvement is higher in the bigger markets as these tend to have a sufficiently large number of individuals with a higher number of total posts.